# From 9-IM Topological Operators to Qualitative Spatial Relations using 3D Selective Nef Complexes and Logic Rules for bodies

Helmi Ben Hmida[1,2], Christophe Cruz[2], Frank Boochs[1], Christophe Nicolle[2]

[1]Laboratoire Le2i, UMR-5158 CNRS, Dep. Informatique IUT Dijon, 7 Boulevard Docteur Petitjean,
BP 17867, 21078 Dijon CEDEX, France
{christophe.cruz, cnicolle}@u-bourgogne.fr

[2]Institut i3mainz, am Fachbereich 1 - Geoinformatik und Vermessung,
Fachhochschule Mainz, Lucy-Hillebrand-Str. 2, 55128 Mainz, Germany
{helmi.benhmida, boochs}@geoinform.fh-mainz.de

Abstract: This paper presents a method to compute automatically topological relations using SWRL rules. The calculation of these rules is based on the definition of a Selective Nef Complexes Nef Polyhedra structure generated from standard Polyhedron. The Selective Nef Complexes is a data model providing a set of binary Boolean operators such as Union, Difference, Intersection and Symmetric difference, and unary operators such as Interior, Closure and Boundary. In this work, these operators are used to compute topological relations between objects defined by the constraints of the 9 Intersection Model (9-IM) from Egenhofer. With the help of these constraints, we defined a procedure to compute the topological relations on Nef polyhedra. These topological relationships are Disjoint, Meets, Contains, Inside, Covers, CoveredBy, Equals and Overlaps, and defined in a top-level ontology with a specific semantic definition on relation such as Transitive, Symmetric, Asymmetric, Functional, Reflexive, and Irreflexive. The results of the computation of topological relationships are stored in an OWL-DL ontology allowing after what to infer on these new relationships between objects. In addition, logic rules based on the Semantic Web Rule Language allows the definition of logic programs that define which topological relationships have to be computed on which kind of objects with specific attributes. For instance, a "Building" that overlaps a "Railway" is a "RailStation".

## 1. INTRODUCTION

Nowadays, qualitative spatial relationships are used in many areas of Computer Science where reasoning about such relationships is fundamental to infer about graphical depiction through logic mechanisms. Such relationships facilitate the access to data by a query processing mechanism that refers to objects and their relationships. Methods for modelling spatial relationships have been compiled in several surveys such as (Galton, 2009) where current models belong to two main categories – connection based model (Randell, et al., 1992), and intersection based one (Egenhofer & Herring, 1990). From a logical point of view, the qualitative models are defined to infer on topological relations without taking into account real geometries. The Open Geospatial Consortium (OGC) has defined a standard nomination to the basic topological relations (Consortium, 2012). From the $\mathbb{R}^3$ space implementation of theses topological relation point of view, (Borrmann, et al., 2009), the octree-based implementation, (Meagher, 1982), and the B-Rep approaches (Lienhardt, 1991) are used to define the spatial operators of a query language. In the octree-approach, Octrees allows the application of recursive algorithms that successively increase the discrete resolution of the spatial objects employed. The B-Rep, approach is used for metric operators such as *mindist, maxdist, isCloserto* and *isFartherfrom.*

From the semantics point of view, the qualitative spatial relations are used to perform inference and to identify inconsistencies on these relations. An ontology based approach is described in (Karmacharya, et al., 2011) and focuses on regions in $\mathbb{R}^2$. The presented approach aims at defining topological relations based on the 9 Intersection Model in $\mathbb{R}^3$, (Ellul & Haklay, 2009), and compute them with the Boolean operators defined by the Nef polyhedra (Granados, et al., 2003). In the actual contribution, the quantitative spatial operators are implemented using built-ins based the Semantic Web Rules Languages (SWRL) which allows the definition of logic program base on Horn-like clauses (Horrocks, et al., 2004). This language is designed to perform logical program on Ontology Web Language (OWL) (Antoniou & Harmelen, 2009). Consequently, the results of these 3D spatial operators may enrich the ontology with spatial relations between the different objects represented via polyhedron. Figure 1 depicts the process sequence for the enrichment of an ontology containing 3D objects.

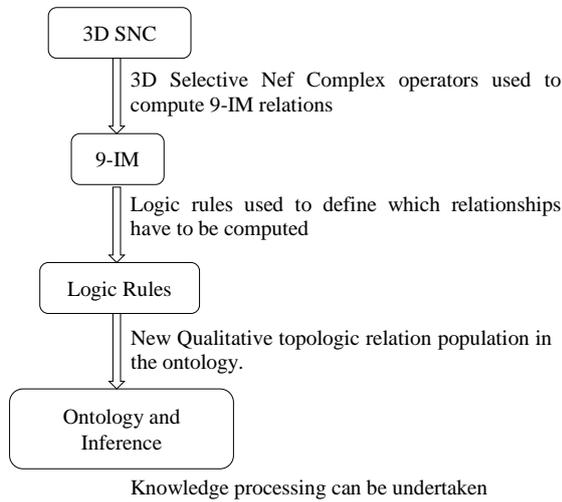

Figure 1: General overview of the process sequence

This paper is divided into 5 sections. Section 2 introduces the technical background on 9-IM, Selective Nef Complex, and logic rules. Section 3 deals with the important elements of the implementation concerning the process sequence. Section 4 highlight the SWRL rules impact on semantic qualification of geometries and finally section 5 concludes the paper.

## 2. BACKGROUND

This section is divided into three sections. The next sub-section focuses on the 9-IM models used to qualify topologic relation. The second sub-section deals with a new data model called Selective Nef Complexes that allows the computation of operators on Nef Polyhedra. These operators are used to verify the topological relation constraints. Finally, the last section deals with the logic aspect and the representation of quantitative topological relations.

### 2.1 Topological relationships

Spatial reasoning is the process that uses spatial theory and artificial intelligence to model and to analyse spatial relations between objects. The standard models are composed by the Simple Feature Relations, The Egenhofer Relations and the RCC8 Relations (Stocker & Sirin, 2009). The Egenhofer Relations are composed of the following relationships: Equals, Disjoint, Meet, Overlap, Covers, Covered by, Inside, Contains (Egenhofer, 2010).

Binary topological relations between two objects, A and B is based upon the intersection of A's interior (A°), boundary ($\delta A$), and exterior ($A^-$) with B's interior (B°), boundary ($\delta B$), and exterior ($B^-$). The 9 intersections between the six objects parts describe a topological relation and can be concisely represented by a 3x3 matrix, called the 9-Intersection Model. The binary relationship R(A,B) between the two objects is then identified by composing all the possible set intersections of the six topological primitives, i.e. $A°\cap B°$, $\delta A\cap B°$, $A^-\cap B°$, $A°\cap \delta B$, $\delta A\cap \delta B$, $A^-\cap \delta B$, $A°\cap B^-$, $\delta A\cap B^-$, $A^-\cap B^-$, and qualifying empty ($\varnothing$) or non-empty ($\neg\varnothing$) intersections. Table 1 shows the 9-IM matrices of the eight topological predicates defined by Egenhofer.

Table 1: The 9-IM matrix

$$R_{(A,B)} = \begin{pmatrix} A° \cap B° & A° \cap \delta B & A° \cap B^- \\ \delta A \cap B° & \delta A \cap \delta B & \delta A \cap B^- \\ A^- \cap B° & A^- \cap \delta B & A^- \cap B^- \end{pmatrix}$$

Table 2: The 6 topological relations between the basic body object, A is the blue box, and B is the red box.

| | | |
|---|---|---|
| A meets B<br>B meets A | | $\begin{pmatrix}\varnothing & \neg\varnothing & *\\ * & * & *\\ * & * & *\end{pmatrix}$ $\begin{pmatrix}\varnothing & * & *\\ \neg\varnothing & * & *\\ * & * & *\end{pmatrix}$ $\begin{pmatrix}\varnothing & * & *\\ * & \neg\varnothing & *\\ * & * & *\end{pmatrix}$ |
| A contains B<br>B inside A | | $\begin{pmatrix}\neg\varnothing & * & *\\ * & * & *\\ \varnothing & \varnothing & *\end{pmatrix}$ |

| A equals B  B equals A | 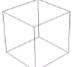 | $\begin{pmatrix} * & \emptyset & \emptyset \\ \emptyset & * & \emptyset \\ \emptyset & \emptyset & * \end{pmatrix}$ |

Table 2 represents the topology in $\mathbb{R}^2$ and $\mathbb{R}^3$ with the 9-IM matrixes for bodies. A basic body object in 3D space is a convex polyhedron that constructed by n (n>2) connected regions (r1, r2, ...,rn). The interior connects and does not contain holes.

## 2.2 Selective Nef Complex

The Selective Nef Complex (SNC) presents a model to define a partition with the labelling of its cells. When the labels are Booleans in order to define the in and out parts, the complex describes a set, a so-called Nef polyhedra (Nef, 1978). In the implementation of Nef polyhedra in 3D (Granados, et al., 2003), they offer a B-rep data structure that is closed under Boolean operations and with all their generality. Starting from halfspaces, it is possible to work with union, intersection, difference, complement, interior, exterior, boundary, closure, and regularization operators. The theory of Nef polyhedra has been developed for arbitrary dimensions. A Nef-polyhedra in dimension $d$ is a point set $P \subseteq \mathbb{R}^d$ generated from a finite number of open half spaces by set complement and set intersection operations and is closed under all Boolean set operations. The implementation in (Granados, et al., 2003) provides functions and operators for the most common ones: complement, union, difference, intersection and symmetric difference. It provides the topological operations interior, closure and boundary. The interior operator deselects all boundary items. The boundary operator deselects all volumes, and the closure operator selects all boundary items.

Table 3: Set of binary and unary operators

| Operators | Syntax |
|---|---|
| Complement | $A^C$ |
| Union | $A \cup B$ |
| Difference | $A \setminus B$ |
| Intersection | $A \cap B$ |
| Symmetric difference | $A \Delta B$ |
| Interior | $I(A)$ |
| Closure | $C(A)$ |
| boundary | $B(A)$ |

## 2.3 Ontology and rules

Ontology is a formal representation of the knowledge through the hierarchy of concepts and the relationships between those concepts. In theory, ontology is a formal, explicit specification of shared conceptualization (Gruber, 1993). Description logics (DLs) (Calvanese, et al., 2001) are a family of knowledge representation languages that can be used to represent knowledge of an application domain in a structured and formally well-understood way. The following example defines a Mother as a Woman which has at least a child type of Person. By inference, it means that every individual type of Women which as at least a relation with a Person and the type of the relation is "hasChild", then this Woman is of kind of Mother.

$$\text{Mother} \equiv \text{Woman} \sqcap \exists \text{hasChild}.\text{Person} \qquad (1)$$

As the Semantic Web technologies matured, the need of incorporating the concepts behind description logic within the ontology languages was realized. It took few generations for the ontology languages defined within Web environment to implement the description language completely. The Web Ontology Language (OWL) (Antoniou & Harmelen, 2009) is intended to be used when the information contained in documents needs to be processed by applications and not by human. The horn logic more commonly known the Horn clauses is a clause with at least one positive literal. It has been used as the base of logic programming and Prolog languages (Sterling, et al., 1986) for years. These languages allow the description of knowledge with predicates. Summarizing, it could be said that ontology defines the data structure of a knowledge base and this knowledge base could be inferred through various inference engines. These inference engines can be perform under Horn logic through Horn-like rules languages. The system of built-ins should also help in the interoperation of SWRL with other Web formalisms by providing an extensible, modular built-ins infrastructure for Semantic Web Languages, Web Services, and Web applications. These built-ins are keys for any external integration, like the integration of the topological operators.

## 2.4 Enrichment of an ontology from Boolean operators

The use of SNC model and its associated Boolean operator allows us to model the topological relationships. In order to combine SWRL rules with topological operators, news built-ins are defined in order to compute the operator. Consequently, the results of the operators can be used to define queries or to enrich the ontology with new topological relationships between two objects. In order to make it possible, two issues appear and have to be solved. First, the semantic definition of the relationships has to be done in the ontology regarding their own properties. Second, the calculation of topological relationships using Boolean operators has to be defined regarding the constraints of the 9-IM model. The following rule specifies that a "Building" defined in the ontology that overlaps a "Railway" defined as well in the ontology, is a "RailStation".

$$\text{Buiding}(?b) \wedge \text{Railway}(?r) \wedge \text{topo:overlaps}(?b, ?r) \rightarrow \text{RailStation}(?b) \quad (2)$$

## 3. IMPLEMENTATION

This section is divided into three sections. The first describes how the Boolean operators are used to compute the 9-IM matrix for a topological relation qualification. The second introduces news relationships in the top-level ontology and its built-in counterparts. These news relationships are specified with a semantic definition. The last section deals with the translation engine which allows the computation of the topological built-ins to enrich the ontology.

### 3.1 Calculation of 9-IM using the SNC Boolean operators

Table 4 presents an overview of the available SNC Boolean operators. Regarding the Table 1 about the 9-IM matrix, only the operators about intersection (A ∩ B), interior (A° equivalent to I(A)), boundary (δA is equivalent to B(A)) and complement ($A^-$ is equivalent to $[I(A) \cup B(A)]^C$ that we will be denoted as E(A)) are necessary. Consequently, the following 9-IM matrix with SNC operators is deduced.

Table 4: The updated 9-IM matrix with SNC operators

$$R_{(A,B)} = \begin{pmatrix} I(A) \cap I(B) & I(A) \cap B(B) & I(A) \cap E(B) \\ B(A) \cap I(B) & B(A) \cap B(B) & B(A) \cap E(B) \\ E(A) \cap I(B) & E(A) \cap B(B) & E(A) \cap E(B) \end{pmatrix}$$

If the results of the nine updated equation is conformed to the expected results then the relation is true. Otherwise the relation is false. Table 5 is an example of the disjoint relation. If one of these equations is false, then the relation between the two objects does not exist.

Table 5:.Example for the disjoint relation

$$R_{(A,B)} = \begin{pmatrix} I(A) \cap I(B) = \emptyset & I(A) \cap B(B) = \emptyset & I(A) \cap E(B) = \neg\emptyset \\ B(A) \cap I(B) = \emptyset & B(A) \cap B(B) = \emptyset & B(A) \cap E(B) = \neg\emptyset \\ E(A) \cap I(B) = \neg\emptyset & E(A) \cap B(B) = \neg\emptyset & E(A) \cap E(B) = \neg\emptyset \end{pmatrix}$$

### 3.2 Definition of topologic relationships in the ontology

Regarding our knowledge base, the top level ontology is created to model the topological relationships. This ontology is used to enrich an existing knowledge base to make it possible to define topological relationships between objects. The next table summarizes for each topological relation, its name in the ontology using the prefix "topo", its semantic characteristics and the new built-in to automatize the computation of relations with the help of SWRL rules. In addition, two inverse relations are defined in the top level ontology. The topo:inside relation is the inverse relation of topo:contains, and the relation topo:covers is the inverse relation of topo:coveredBy.

Table 6: Definition of the topological relationships and its semantics

| Topologic relations | Pro-perty | Characteris-tics | SWRL built-ins |
|---|---|---|---|
| Meets | topo: meets | Symmetric Irreflexive | swrl_topo:meets (?x, ?y) |
| Inside | topo: inside | Transitive Asymmetric Irreflexive | swrl_topo:inside (?x, ?y) |
| Equals | topo: equals | Transitive Symmetric Reflexive | swrl_topo:equals (?x, ?y) |

### 3.3 Translation engine

The translation engine allows the computation of spatial SWRL rules which can also be in form of queries. It interprets the statements in order to parse the spatial components. Once the spatial components are parsed, they are computed through relevant spatial functions and operations by the translation engine through the operations provided at the SNC level. The results are populated in the knowledge base, thus making it spatially rich. After that, the spatial statements are translated to standard ones for the executions through their respective engines. With the inference engine, the enrichment and the population of the ontology through the results of the inference process is eventually stored in the knowledge base, Figure 2.

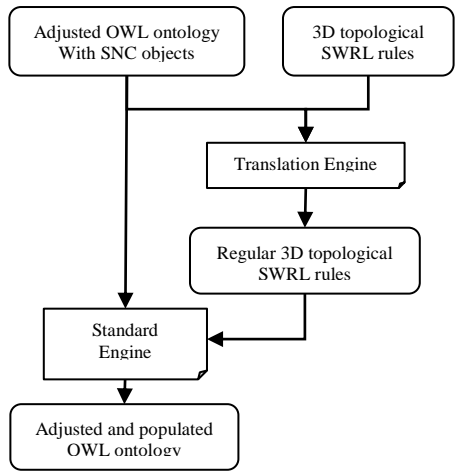

Figure 2: The translation engine that process rules with topological built-ins.

## 4. SWRL RULES IMPACT ON SEMANTIC QUALIFICATION

To highlight the utilisability of the presented approach, we decide to extend the research by making a step forward from the qualification of the spatial relation semantically to the extension of the semantic rules and query language creating a 3D Semantic Spatial Qualification platform (*3DSQ*). Such an improvement will support the inference on 3D spatial knowledge and will allow finally querying spatial knowledge base. To do, a Java prototype demonstrating the applicability of the presented concept was developed. It ensure the interaction between the users, the OWL ontology and the Qualification engine from one side and maintain an interactive visualisation of the qualified Spatial Relation from another side.

Add to its ability to process Spatial Data in our case, *3DSQ* platform guarantee a common understanding of Spatial domain between Human and machines via ensuring the Semantic inference and queries using Spatial knowledge. The declaration of the spatial built-Ins in our cases respects the standard nomination suggested by Egenhofer (Egenhofer, 2010). As convention, each Topological Built-Ins began with the prefix "Swrlb_Topo" where the first syllable state that it presents a complex Built-Ins while the second one highlight the type of the Built-Ins, "Topo" in our case. Finally, the type of spatial topological predicate, "Inside" for example, will validate such a relation. In such case, the scene in question and the spatial qualification results are presented in different colours depending on the nature of objects.

To prove it, an IFC architectural scene containing just geometric elements was populated in the ontology. Furthermore, spatial relations between populated geometries are qualified via the *3DSQ*. In the next, two examples reflecting the main languages acting with the semantic web one which are the SWRL and SQWRL (Semantic Query-Enhanced Web Rule Language) one will be highlighted.

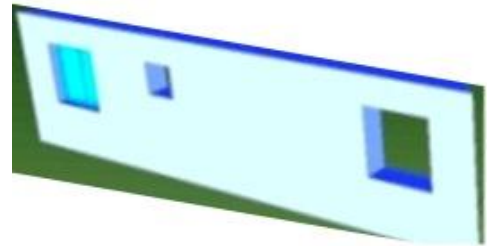

Figure 3: Example of an Inside Relationship (Blue Sky elements)

First, the rule 3 shows an example able to extract from the knowledge base all the elements "Inside" walls individuals which can be qualified later on if respecting certain characteristics as windows for example, Figure 3.

Wall(?x) ∧ Geometry(?y) ∧ swrlb_Topo:Inside  (3)
(?x,?y) ∧ haslength(?y,?l) ∧ swrlb:LessThan
(?l,2) → Windows (?y)

Second, as we have already selected a qualitative manner based on semantic knowledge to define spatial operators, SQWRL (**S**emantic **Q**uery-Enhanced **W**eb **R**ule **L**anguage) language can be used as a query language to query the knowledge base. The next equation (rule 4) is an example of a query that select all "Meets" spatial elements with the element "*BldgElem__113_BBox*" in the current knowledge base, Figure 4.

Geometry(BldgElem__113) ∧ Geometry (?y)  (4)
∧ Swrlb_Topo: Meets (BldgElem__113, ?y)
→ sqwrl:select(?y)

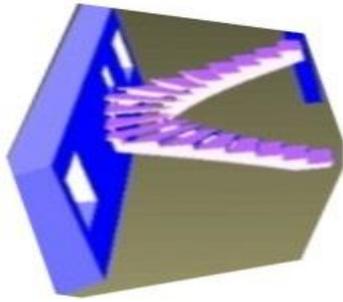

Figure 4: Example of SQWRL rule result

## 5. CONCLUSION

This paper presents a semantic method to compute automatically topological relations using OWL ontology and SWRL rules. The calculation of these rules is based on the definition of Nef Polyhedra which can be generated automatically from standard Polyhedron. A prototype is being developed using the library CGAL (http://www.cgal.org). Some simplification will be undertaken regarding the 9-IM computation of each topological relationship in order to reduce the calculation volume. Future work on topological relation qualification will be mainly focus on semantic qualification and inferences (Boley, et al., 2001) and depicted in the next generation of SWRL topologic rule. This can also be done by a composition of relations, meet ∘ contains ⊑ Disjoint.

$$\text{meet (?a, ?b)} \wedge \text{contains(?a, ?c)} \rightarrow \text{disjoint (?a, ?c)} \quad (4)$$